\documentclass{icbo}

\usepackage{url}
\usepackage{graphicx}
\usepackage{tawny}
\usepackage{xspace}
\usepackage{hyperref}
\usepackage{float}
\usepackage{subcaption}
\usepackage[export]{adjustbox}

\lstMakeShortInline[style=tawnystyle]| \lstnewenvironment{tcode}[1][]
{\lstset{style=tawnystyle,frame=shadowbox,basicstyle=\footnotesize\ttfamily,#1}}{}

\newcommand{\tawny}{Tawny-OWL\xspace}
\newcommand{\protege}{Prot\'eg\'e\xspace}

\begin{document}

\title[Readable and Editable Ontologies]{User and Developer
  Interaction with Editable and Readable Ontologies}

\author[A. Blfgeh \textit{et~al}]{Aisha Blfgeh\,$^{1,2}$\footnote{To
    whom correspondence should be addressed:
    \href{mailto:a.blfgheh1@ncl.ac.uk}{a.blfgeh1@newcastle.ac.uk} or
    \href{mailto:abelfaqeeh@kau.edu.sa}{abelfaqeeh@kau.edu.sa}} and
  Phillip Lord\,$^{1}$}
\address{$^{1}$School of Computing Science, Newcastle University, UK\\
  $^{2}$Faculty of Computing and Information Technology, King
  Abdulaziz University, Jeddah, Saudi Arabia}

\maketitle

\begin{abstract}

  The process of building ontologies is a difficult task that involves
  collaboration between ontology developers and domain experts and
  requires an ongoing interaction between then.  This collaboration is
  made more difficult, because they tend to use different tool sets,
  which can hamper this interaction. In this paper, we propose to
  decrease this distance between domain experts and ontology
  developers by creating more readable forms of ontologies, and
  further to enable editing in normal office environments.

  Building on a programmatic ontology development environment, such as
  \tawny, we are now able to generate these readable/editable from
  the raw ontological source and its embedded comments.  We have this
  translation to HTML for reading; this environment provides rich
  hyperlinking as well as active features such as hiding the source
  code in favour of comments. We are now working on translation to a
  Word document that also enables editing.
  
  Taken together this should provide a significant new route for
  collaboration between the ontologist and domain specialist.

\end{abstract}

\section{Introduction}
\label{sec:introduction}

Ontologies are wide-spread in the field of biology and biomedicine, as
they facilitate the management of knowledge and the integration of
information, as in the \emph{Semantic}
Web~\citep{Bermejo2007}. Additionally, biological data are not only
heterogeneous but also require complex domain knowledge to be dealt
with~\citep{Stevens2000a}. Therefore, ontologies are useful models for
representing this complex knowledge that is potentially changing and
are also widely used in biomedicine, examples being the GO (Gene
Ontology)~\citep{Ashburner2000}, SNOMED (Systematized Nomenclature of
Medicine)~\citep{Organisation2016}.

However, building an ontology is a challenging task due to the use of
languages with a sophisticated formalism (such as OWL), especially when
combined with a complex domain such as biology or medicine. Normally
ontologies are built as a collaboration between domain specialists who
have the knowledge of the domain and ontology developers who know how
to structure and represent the knowledge; they have to work together
to construct a robust and accurate ontology. Often, community
involvement during the process of building ontologies using meetings,
focus groups and the like is very important~\citep{Mankovskii2009}, as
in GO where biological community involvement is important for
successful uptake~\citep{Bada2004}. In addition, ~\citet{Bult2011}
state that the development of Protein Ontology requires wider range of
involvement to include other users and developers of the associated
ontologies (such as GO) to ensure consistent architecture of the
ontology.

Biologists represent, manipulate and share their data in a
wide-variety of tools such as Microsoft Excel spreadsheets and Word
documents. Unfortunately, these environments are far removed from the
formal structured representation of the ontology development
environments with which the ontologists work to build ontologies. As a
result of this difference in tools it is unclear how we can bridge the
gap between the the two groups; this would be useful to facilitate the
interaction between domain specialists and ontologists and help to
make more convenient for both sides to read and/or manipulate the
ontology.

Ontology development environments are designed to produce formal
structured representation of any domain. Either using GUI software
such as \protege~\footnote{\url{http://protege.stanford.edu/}} or a
textual programmatic environment such in \tawny~\citep{Lord2013a}. The
next section describes these tools in more details.

\section{Building ontologies}
\label{sec:buildont}

There are various tools for constructing and developing ontologies
with a variety of user interfaces and environments. The most popular
is \protege which is an open-source tool that provides a user
interface to develop and construct ontologies of any domain. It has
been widely used for developing ontologies due to the variety of
plug-ins and frameworks~\citep{Noy2003}. \protege provides an easy
interface for editing, visualisation and validation of ontologies as
well as a useful tool for managing large
ontologies~\citep{Horridge2011}.

Conversely, Tawny-OWL is a textual interface for developing ontologies
in a fully programmatic manner~\citep{Warrender2015}. This provides a
convienient and readable syntax which can be edited directly using an
IDE or text editor; in this style of ontology development, the
ontologist ceases to mainpulate an OWL representation directly, and
instead develops the ontology as programmatic source code. In contrast
to developing ontologies in OWL, the ontologist can introduce new
abstractions and syntax as they choose, whether for general use or
specifically for a single ontology. An OWL version of the ontology
can then be generated as required. It has been implemented in Clojure,
which is a dialect of lisp and runs on the Java Virtual
Machine~\citep{Lord2013a}. Like \protege, it also wraps the
OWL-API~\citep{Mankovskii2009} which performs much of the actual work,
including interaction with reasoners, serialisation and so forth.

Recently, we have developed tolAPC ontology using a new
document-centric approach by including an Excel spreadsheet directly
in the development pipeline. The spreadsheet contains all knowledge
for the ontology which has been created and maintained by a
biologist. Meanwhile, we design the ontology patterns using \tawny,
then generate the axioms by extracting data from the spreadsheet using
Clojure. Thus, \tawny contains the spreadsheet as a part of the source
code; which can be freely updated and the ontology regenerated when
needed. Hence, it remains as a part of the ontology development
process~\citep{BlfgehWHL16}.

In this approach, the Excel spreadsheet is totally developed by
biologists; this has a significant advantage because it is a tool
which they are familiar with and find convenient. However, we cannot
ensure that the programmatic transformation of the values in the
spreadsheet to the final ontology conforms with the domain specialists
understanding, without the biologists reading and interacting with
source code. Therefore, next we will discuss the probabilities of
making this ontological source more readable by the specialists.

\section{Multilingual ontologies}
\label{sec:multilang}

The first and most obvious mechanism for increasing ontology
readability is to enable users to read and write the ontology using
their native language. Internationalisation technologies are
widespread and enable support for multiple languages for applications
with a graphical user interface.

We next consider how we can enable support for multiple languages in
textual user interface such as \tawny, giving the ontologist the
ability to use their own native language for all parts of the
development process.

\begin{figure*}[!htpb]
 \centering
 \includegraphics[scale=0.7]{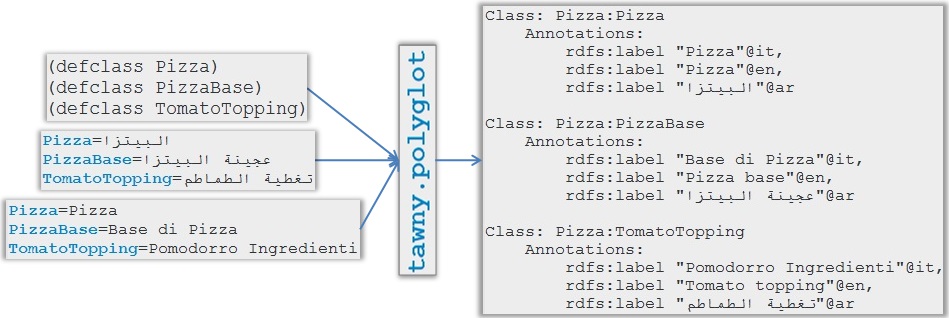}
 \caption{Polyglot library in \tawny.}
 \label{fig:polyglot}
\end{figure*}

The first option is using polyglot library. This part of the system
mimics a fairly standard technique for internationalisation of
programmatic code; the ontology is developed with a set of
programmatic labels which are then referenced in a language, or locale
bundle with an appropriate translation. In the case of \tawny, this
translation appears as |rdfs:label| annotations on the ontology
entities (classes, properties etc). This overall process is shown in
Figure~\ref{fig:polyglot}, placing Italian and Arabic language
translations onto the pizza ontology.

\begin{figure*}
\centering
\begin{subfigure}[b]{.49\linewidth}
\includegraphics[width=\linewidth]{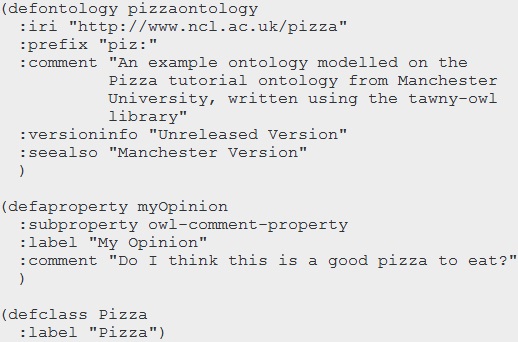}
\caption{English Pizza Ontology}\label{fig:mouse}
\end{subfigure}

\begin{subfigure}[b]{.45\linewidth}
\includegraphics[width=\linewidth]{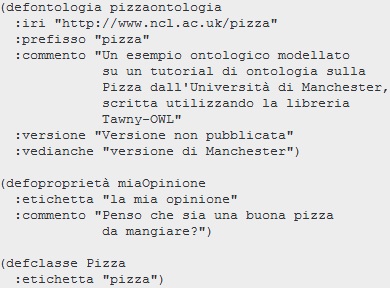}
\caption{Italian Pizza Ontology}\label{fig:gull}
\end{subfigure}
\begin{subfigure}[b]{.48\linewidth}
\includegraphics[width=\linewidth]{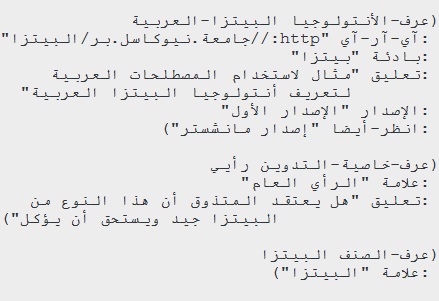}
\caption{Arabic Pizza Ontology}\label{fig:tiger}
\end{subfigure}
\caption{Multilingual Pizza ontology}
\label{fig:multilang}
\end{figure*}

While this may enable internationalisation for users of the ontology,
it does not change the English-centric editing environment. We would
wish, instead, to internationalise the entire source code of the
ontology.  This will make the entire ontology more comprehensible and
readable for all developers who communicate in Italian and/or
Arabic. This is fully supported with a full conversion of the
environment using the multilingual feature of \tawny as in
Figure~\ref{fig:multilang}, which shows the English, Italian and
Arabic version of the pizza
ontology~\citep{LordPizza2013}.
The latter of these is a right-to-left alphabet, and we can use the
IDE to change the direction that \tawny code is rendered in. This
demonstrates the capability of \tawny to adapt with any language. The
next language to be implemented will be French.

These multilingual environments are advantageous for being readable
and comprehensible by users when using their own language.  This still
leaves us in a programming environment, which is an environment
unlikely to be familiar or comfortable to the most domain
users. Moreover the ontology lacks a narrative structure, which means
that it cannot be read in a literate fashion. We consider how to
enable this in the next section.

\section{Literate ontologies}
\label{sec:literate}

The term literate programming was invented
by~\citep{knuth1992literate} where the program is treated as a piece
of literature rather than a program. The main idea in this paradigm is
to insert text along with code and the program will also be its own
documentation. The intentionality here is that the program should
become easier to understand and, conversely, that the documentation is
less likely to become out-of-date, as it is maintained in the same
place.
 
As \tawny is a fully programmatic environment, we can add comments
freely, along with any additional mark-up that we wish. This enables
us to produce different representation of the ontology.

\begin{figure*}[!htpb]
 \centering
 \includegraphics[scale=0.65, frame]{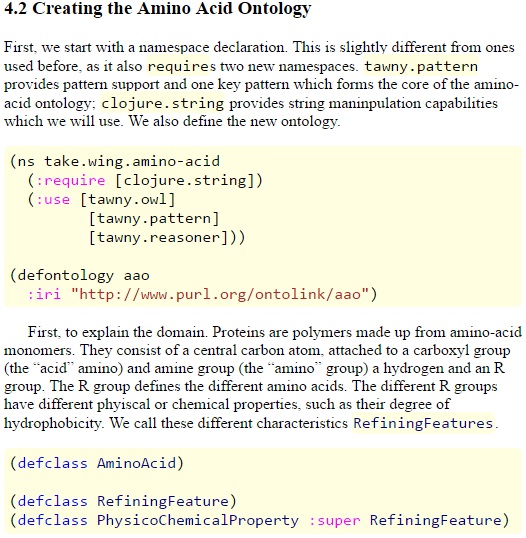}
 \caption{Literate Amino Acid ontology in HTML representation.}
 \label{fig:amino-acid}
\end{figure*}

We have previously discussed two examples of literate ontologies: the
first is the Amino Acid Ontology, taken from a previous ICBO2015
tutorial about \tawny~\footnote{\url
  {http://homepages.cs.ncl.ac.uk/phillip.lord/take-wing/take_wing.html}},
while the second is a version of the Karyotype
ontology~\citep{LordW15}. In both cases, they have been produced using
the \tawny source code, with markup in the comments being interpreted
using a markup processing tool.  Figure~\ref{fig:amino-acid} shows a
snippet from the literate ontology Amino Acids as a webpage. The
result appears as a normal web page, with syntax highlighting for the
source code.

Literate ontologies can be represented in different forms; using the
various techniques for converting the markup text into different
formalisms; webpages for example. Representing the ontology as an HTML
webpage gives us the ability to navigate and browse the documentation
either in order (section by section) or with a navigation facility
(jumping between sections). It is also possible to hide or expose the
``source'' sections, leaving the reader to see just the documentation
as appropriate.  From the developer perspective, while the reader may
still not be able to see the axiomatization in this way, the comments
that they have checked are embedded directly next to the code which is
an interpretation of them.

It is interesting to enable specialists to read and navigate through
the ontology and its documentation. However, with HTML there are no
editing facilities to modify and update the ontology. Therefore,
rather than using HTML, we have also investigated the possibility to
turn the whole ontology into a Word document, an environment which can
also be modified, changed or updated. Now biologists and domain
specialists are placed in an environment in which they can freely
provide feedback on an existing ontology simply by interacting with
Word documents.

\section{Discussion}

In this paper, we have described our approach to the translation of
ontologies into a form that domain users can interact with more
naturally.

We have shown that it is possible to translate a textual environment
like \tawny into another human language, or indeed a different script,
including right-to-left text. To our knowledge, this is the first
ontology editing environment with such textual and syntactic
flexibility.  Despite the fact that the multilingual ontologies
approach is less relevant for scientific ontologies, it is already
applied by some means in some cases of terminologies.  For example,
the use of \texttt{some} and \texttt{only} in \tawny rather than using
\emph{universal} and \emph{existential} notations which implies the
agreement for using alternative language for ontology development.

Further than this, however, we also translate the ontological source
code into alternative visualisations such as HTML and Word documents
which map directly back to the source, but which can differ from it:
for instance, by enabling hyperlinks, adding section links and hide
source code in favour of commentary. Especially with a Word document,
this should enable a novel mechanism for interacting with an ontology:
users can see and edit comments, with change tracking switched on, and
use this as mechanism for feeding back to the ontology developer.

Using this approach, of course, only enables us to visualise
ontologies developed using \tawny. While a migration path is
provided~\citep{Warrender2015}, a whole-sale switch to \tawny is not
effort-free. We note, however, that many ontologies are developed
partly in \protege and partly using OWL generated from other sources; a
secondarly migratory path would be to use \tawny for these sections.

We still need to evaluate this kind of interaction rigorously. For
this, we are proposing a focus group test which will be include
specialists participants to read the document of the ontology and
provide their opinion about it and whether they prefer to update any
terminologies according to their expertise.


We are not proposing that Word documents will be directly used by
domain specialists for editing ontologies. We expect that an
ontologist will be involved with incorporating changes suggested back
to the domain user; in this sense, we are using a Word document as an
\emph{intermediate representation}~\citep{Rector2001}. Our hope is
that the reviewing features of Word should, however, enable us to
provide a rich environment to support the ontologist in this
process. Taken together, these should provide us with a significantly
enhance process for the knowledge capture, ontology development and
refinement from the process that we currently have.

\section*{Acknowledgements}

Thanks to Newcastle university for supporting this research. Also, thanks
to King Abdulaziz University, Jeddah, Saudi Arabia for funding and
supporting the study.

\bibliographystyle{natbib}

\bibliography{aisha_blfgeh_icbo_2017}

\end{document}